\documentclass[conference]{IEEEtran}
\IEEEoverridecommandlockouts
\usepackage{mathtools}
\usepackage{subcaption}
\usepackage{cite}
\usepackage{amsmath,amssymb,amsfonts}
\usepackage{algorithmic}
\usepackage{graphicx}
\usepackage{textcomp}
\usepackage{xcolor}
\usepackage{url}
\def\BibTeX{{\rm B\kern-.05em{\sc i\kern-.025em b}\kern-.08em
    T\kern-.1667em\lower.7ex\hbox{E}\kern-.125emX}}
\begin{document}

\title{ Towards Interpretable Multilingual Detection of Hate Speech against Immigrants and Women in Twitter at SemEval-2019 Task 5 }

\author{\IEEEauthorblockN{ Alvi Md Ishmam}
\IEEEauthorblockA{\textit{Department of Computer Science and Engineering} \\
\textit{Bangladesh University of Engineering and Technology}\\
Dhaka, Bangladesh \\
1305092.am@ugrad.cse.buet.ac.bd}
}

\maketitle

\begin{abstract}

This paper describes our techniques to detect hate speech against women and immigrants on Twitter in multilingual contexts, particularly in English and Spanish. The challenge was designed by SemEval-2019 Task 5, where the participants need to design algorithms to detect hate speech in English and Spanish language with a given target (e.g., women or immigrants). Here, we have developed two deep neural networks (Bidirectional Gated Recurrent Unit (GRU), Character-level Convolutional Neural Network (CNN)), and one machine learning model by exploiting the linguistic features. Our proposed model obtained 57 and 75 F1 scores for Task A in English and Spanish language respectively. For Task B, the F1 scores are 67 for English and 75.33 for Spanish. In the case of task A (Spanish) and task B (both English and Spanish), the F1 scores are improved by 2, 10, and 5 points respectively. Besides, we present visually interpretable models that can address generalizability issues of the custom-designed machine learning architecture by investigating the annotated dataset.

\end{abstract}


\section{Introduction}
The means of mass communication have been influenced drastically since the introduction of Social Networking Sites (SNS) allows anyone to express thoughts and opinions more effectively to larger communities. However, exploiting these services from SNS, haters spread hateful content towards particular individuals or groups (e.g., women, religious minorities, refugees). Therefore, the automatic identification of the hateful speech draws the attention of social network companies, law enforcement agencies, and researchers. The interested stakeholders put collaborative efforts to develop open-source datasets for multilingual hate speech detection and semantic evaluation. In this paper, we investigated the dataset of SemEval-2019 Task 5 (hatEval) \footnote{\url{https://competitions.codalab.org/competitions/19935}}, where participants should algorithmically detect hate speech, a two-class classification problem, where systems have to predict whether a tweet in English or Spanish with a given target (women or immigrants) is hateful or not hateful (Task A). Moreover, the system needs to detect whether a hate speech is directed at a specific person or a group of individuals (Task B).

In this paper, we have three contributions. First, we employed three state-of-the-art machine learning and deep learning algorithms, along with the analysis of the linguistic features of raw tweets. Our proposed model outperforms the existing baseline F1-score by at most 10 points for Task B of the hatEval competition. Secondly, we have investigated the interpretability of the models using Local Interpretable Model-Agnostic Explanations (LIME) \cite{LIME} to understand the performance of the algorithms. Finally, a quantifiable error analysis gives a clear insight into why an apparently good model is underperforming in tests and real environments.

\section{Related Works}

Researchers work on hate or offensive speech detection in multilingual contexts in web blogs and social media. SemEval has a series of evaluation where the latest publicly available datasets are used for semantic evaluation \cite{ABARUAH, Atalaya, CiTIUS, Fermi, fersini1,MineriaUNAM, MITRE, NLPDove, guir, Duluth}. In these system papers, the authors focus on maximizing the competition evaluation matrices (macro average F1 score) by implementing various machine learning and deep learning architectures. However, these papers do not address how linguistic features are used in models except for little quantitative analysis of the dataset. In a nutshell,  few studies are interested in the interpretation of the models and why these models do not perform well enough in the real environment despite promising performance in training sets. In our study, apart from the performance of the system, we have investigated both train and test datasets for identifying the reasons why a model performs better than the other by providing visual interpretations of features. Moreover, we have provided quantitative and linguistic analysis to present the inconsistencies between the train and test dataset.

\section{Dataset Description}   

The hatEval dataset is composed of 19,600 tweets, 13,000 tweets for English, and 6,500 for Spanish. The English dataset is composed of 9000 tweets for the training, 1000 for the development, and 3000 for the test set. The Spanish dataset is half the size of English with 4500 tweets for the train, 500 for development, and 1500 for the test \cite{basile2019semeval}. 

The tweets were accumulated from July to September 2018. The training dataset against women was developed in the context of two previously challenges of misogyny identification \cite{fersini1,fersini2018overview}. To rigorous understanding in hate speech, offensiveness, and stance, keyword-driven approach along with neural keywords, derogatory words against the targets, and highly polarized hashtags were used to annotate the datatset. For instance, the frequent keywords used for the English language are \textit{migrant, refugee, \#buildthatwall, bitch}, etc.

\section{Task Description}

HatEval was the number 5 task organized by the SemEval-2019 \cite{basile2019semeval}.  The organizers divided the multilingual tasks into two parts in both English and Spanish language for a total of four sub-task evaluations.

\subsection{Subtask A:} The first subtask was a binary classification of hate speech (HS) against women or immigrants both in English or Spanish language. In this binary classification task, the classes are labeled as hate speech (HS=1) and non-hate speech (HS=0).

\subsection{Sutask B:} The second subtask can be considered as a form of rigorous understanding of hate speech. After considering a tweet as hate speech (HS=1),  participants need to determine whether the tweet is incitement against an individual or a group of people (e.g., women or immigrants). Therefore, tweets representing generic and individual hate are annotated as targeted hate speech (TR=1) or not (TR=0). Moreover, the hateful speech (HS=1) is classified as aggressive (AG=1) or non-aggressive (AG=0) in various forms such as violent acts against the target \cite{sanguinetti2018italian}.

\section{Evaluation Methodology and Baseline}

The evaluation process is different for both subtask A and subtask B to avail more fine-grained scores. The submission was ranked depending on the macro averaged F1 score. The following strategies were employed for both subtasks.

\textbf{Subtask A:} The submission was evaluated based on standard evaluation matrices \textit{precision, recall, F1-score, Accuracy}. The matrices are computed as follows:

$$ Precision = \frac{number of correctly predicted instances}{number of predicted labels}$$

$$ Recall = \frac{number of correctly predicted labels}{number labels in the gold standard} $$

$$ F1-score = \frac{2 \times Precision \times Recall}{Precision + Recall} $$
$$Accuracy = \frac{number of correctly predicted instances}{total number of instances} $$

\textbf{Subtask B:} The evaluation of models submitted for subtask B was determined based on two criteria: 1) partial match 2) exact match. The official metric used for subtask B was the Exact Match Ratio (EMR), which is the proportion of tweets that are labeled correctly for all categories (hate speech, targeting, and aggression). The partial match was calculated depending on each category to be predicted (HS, TR, AG) independently from others using standard evaluation matrices. The exact match was reported in terms of the average macro average score calculated as follows: 

 $$F1-score = \frac{\splitfrac{F1-score(HS) + F1-score(TR)}{+ F1-score(AG)}}{3}$$

\section{ Proposed Model} Before feeding the tweets to the models, we processed the raw tweets to remove bad characters, noisy words or characters, punctuations. The cleaned text are stemmed after tokenization. The snowball stemmer in the nltk package was used for the English and Spanish languages. Then, we proposed one machine learning and two deep learning models for these classifications problems. The models are :                                                                                                                                      

\begin{itemize}
    \item Bidirectional Gated Recurrent Unit (GRU) based Model
    \item Character Level Convolutional Neural Network (CNN)
    \item Feature Engineering with linguistic features with Linear SVC
\end{itemize}{}

\subsection{Bidirectional GRU:} 
\begin{figure}[h!]
    \centering
    \includegraphics[width=\linewidth]{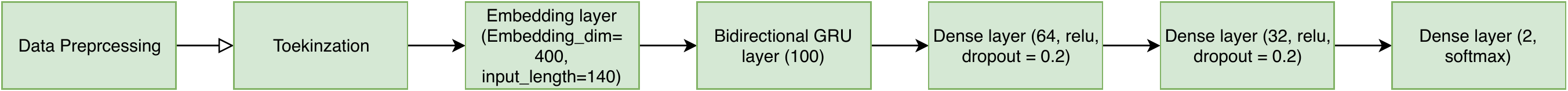}
    \caption{GRU architecture}
    \label{fig:GRU architecture}
\end{figure}

In Fig. \ref{fig:GRU architecture}, we present the architecture of the Gated Recurrent Unit (GRU) based neural network. We go for GRU instead of Long-Short Term Memory (LSTM) based network for faster training and achieve good performance over small dataset i.e., GRU has a better ability to generalize and less tendency to overfit on small datasets \cite{GRU}. To implement word embedding, firstly, the preprocessed tweets are tokenized using Keras tokenizer API \footnote{\url{https://keras.io/preprocessing/text/}} and converted into a list of sequences. The maximum sequence length is 140 since no tweet is longer than 140 characters when the dataset was developed. Then, the sequences are fed into an embedding layer with an output dimension of 400. The output of the embedding dimension (140, 400) is then fed into a spatital1D dropout layer of Keras with a dropout rate of 0.2. The layer is used as regularization to avoid overfitting. The bidirectional GRU layers with 100 hidden layers are followed by two dense layers with a dimension of 64 and 32 respectively having a dropout rate of 0.2. The activation function is `relu'. The final output is a binary classification of tweets (HS vs Non-HS, TR vs Non-TR, or AG vs Non-AG). Therefore, the final output is fed into a dense layer having `softmax' activation function with two dimensions. We performed training in batches of size 32 by using `adam' optimizer. We evaluated the experiment ten times for each configuration and reported the average performance. In each iteration, we reserved 10\% of our data for testing, and the rest was divided into 90\% train and 10\% validation set for the fine-tuning of the hyperparameters (dropout rate, hidden layers).

The macro average F1 score of subtask A and subtask B for the Bidirectional GRU model are shown in Table \ref{tab:Macro average f1 score for two subtask}. Here, EN and ES stand for English, and Spanish respectively. HS, TR, AG represent \textit{Hate Speech}, \textit{Targeted Hate}, and \textit{Aggressive Speech} respectively. For instance, in Table \ref{tab:Macro average f1 score for two subtask}, EN\_HS means hate speech in English.

\begin{table}{}
\begin{minipage}{0.4\linewidth}
        \centering
\caption{Macro average f1 score for  subtask A and B for Bidirectional GRU.}
\label{tab:Macro average f1 score for two subtask}
\begin{tabular}{ | c | c | }
\hline
 Class & F1-score\\
 \hline
 EN\_HS &  47\\
 \hline
 EN\_TR &  83 \\
 \hline
 EN\_AG &  61 \\   
  \hline
 ES\_HS &  72 \\ 
  \hline
 ES\_TR &  88 \\ 
  \hline
 ES\_AG & 62\\ 
  \hline
\end{tabular}
\end{minipage}
\quad
\begin{minipage}{0.4\linewidth}
        \centering
\caption{Macro average f1 score for two subtask using Character CNN}
\label{tab:Macro average f1 score for two subtask in CharCNN}
\begin{tabular}{ |c |c |}

\hline
 Class & F1-score\\
 \hline
 EN\_HS &  46\\ 
 \hline
 EN\_TR &  79 \\  
 \hline
 EN\_AG &  57 \\   
 \hline
 ES\_HS &  64 \\ 
 \hline
 ES\_TR &  80 \\ 
 \hline
 ES\_AG & 65\\ 
 \hline
\end{tabular}
\end{minipage}
\end{table}
\quad

\subsection{Character Level Convolutional Neural Network (CNN)}
People use misspelled words on Twitter, and in some cases misspelled or noisy words are used more frequently for expressing similar contextual meaning. In tweets, word shorthands e.g., `FYI’ instead of `for your information’ or repeating characters e.g., `yaaayyyyyy’ or emoticons are used to express the emotion. These are not general English words, and these usages may differ from person to person. For example, `yaaaayyyyy’, `yaayyy’, `yeah' express a similar meaning though the first two are misspelled words. Therefore, we go for character-level CNN for rather than words or token level CNN. 

In this approach, we came up with the idea of embedding the characters of the words and then pass it through the CNN network. The model accepts a sequence of encoded characters as input. The encoding is done by prescribing an alphabet of size m for the input language, and then quantize each character using the 1-of-m encoding (or ``one-hot” encoding). Then, the sequence of characters is transformed into a sequence of such m sized vectors with fixed lengths. Any character sequence exceeding maximum length is ignored, and any characters that are not in the alphabet including blank characters is quantized as all-zero vectors. The character quantization order is backward so that the latest reading on characters is always placed near the beginning of the output, making it easy for fully connected layers to associate weights with the latest reading. The alphabet used in all of our models consists of 70 characters, including 26 English letters, 10 digits, 33 other characters, and the new line character. The non-space characters are: abcdefghijklmnopqrstuvwxyz0123456789 |-,;.!?:’’’/\textbackslash \_@\%ˆ\&*˜‘+-=<>()[]{}\#\$. 

\begin{figure}[h!]
    \centering
    \includegraphics[width=\linewidth]{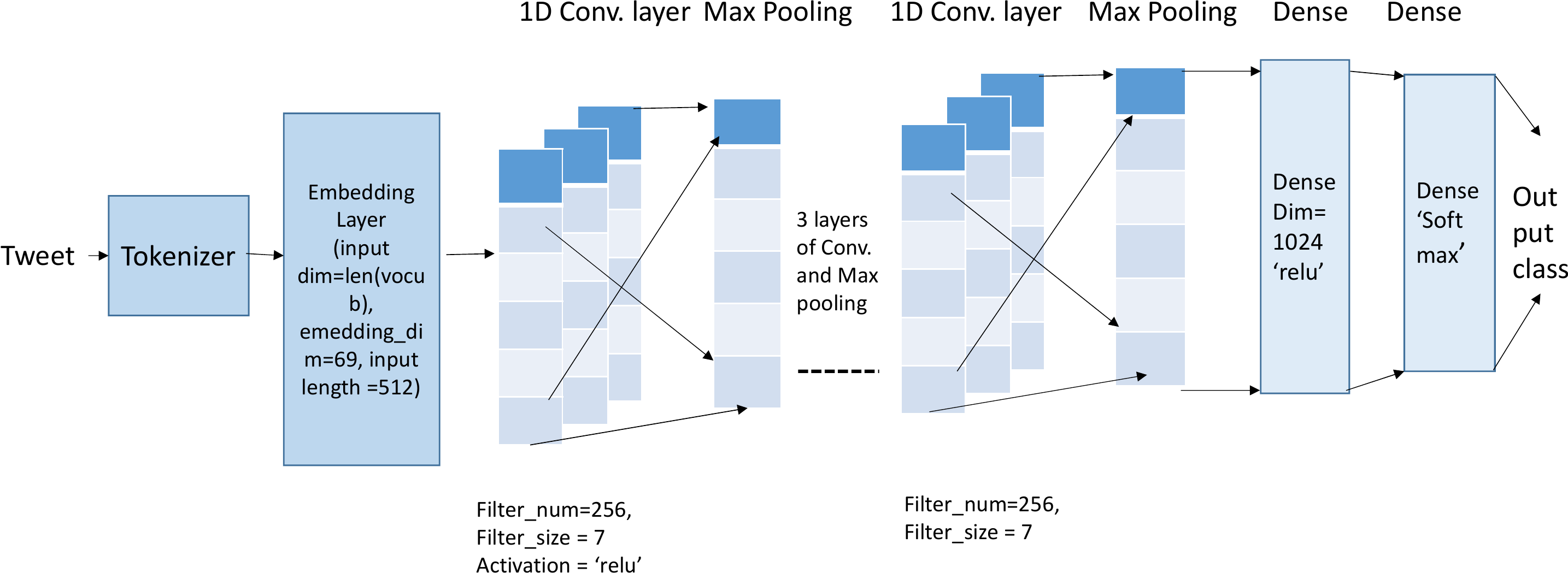}
    \caption{CNN architecture}
    \label{fig:CNN architecture}
\end{figure}

The embeddings are then passed to the described CNN network. We used three 1D convolutional layers \footnote{\url{https://keras.io/layers/convolutional/}} with 256 filters, and the kernel size is 7. Each layer is followed by a max-pooling layer with pooling size 3. We also insert 2 dropout modules in between the 3 fully-connected layers to regularize. They have a dropout probability of 0.2. The activation function for the convolutional layer, as well as the dense layer, is `relu'. We used `adam' optimizer for regularization and the loss function is `categorical\_crossentropy'. In Fig. \ref{fig:CNN architecture}, the CNN architecture is shown. The macro averaged F1-score is given in Table \ref{tab:Macro average f1 score for two subtask in CharCNN}.

\subsection{Feature Engineering with State of the Art Machine Learning}

The character level CNN performs better in the case of the Spanish language (Targeted and Aggressive) language. In other cases, the model does not perform better as compared to the GRU model. Then, we plan to investigate the linguistic along with the quantitative features from tweets. The features are the following :

\begin{itemize}
    \item N-gram features each token weighted by its Term Frequency - Inverse Document Frequency (TF - IDF)
    \item Text quality: Flesch-Kincaid Reading Ease and Flesch
Grade Level Score.

    \item Syllable count per tweet, word count per tweet, character count per tweet, length per tweet, number of capitals per tweets, number of unique words per tweet, word density per tweet
    
    \item Hashtag (\#) count per tweet, annotation count (\@) per tweet
    
    \item Positive, negative, neutral sentiment score word per word from a tweet
    \end{itemize}

\subsubsection{N gram features each token weighted by its TF - IDF:} The keywords of this feature are N-gram token and TF-IDF. For example,
bi-gram means consecutive two tokens. In our case, we have
taken N-gram, which means N (3 or 5 tokens) consecutive
tokens to be considered. The second keyword is Term
Frequency (TF) and Inverse Document Frequency (IDF). The formal definition of the two parameters are –

$$TF(t) = \frac{\#\_of\_times\_term\_t_appears\_in\_document}{Total\_number\_of\_terms\_in\_the\_document}$$

$$IDF(t) = \log \frac{Total\_number\_of\_documents}{Number\_of\_documents\_with\_term\_t\_in\_it} $$

We have calculated the TF-IDF using the following formula.
$$TF-IDF(t) = TF(t) \times IDF(t)$$

\begin{figure}[!h]

 \includegraphics[width=\linewidth ]{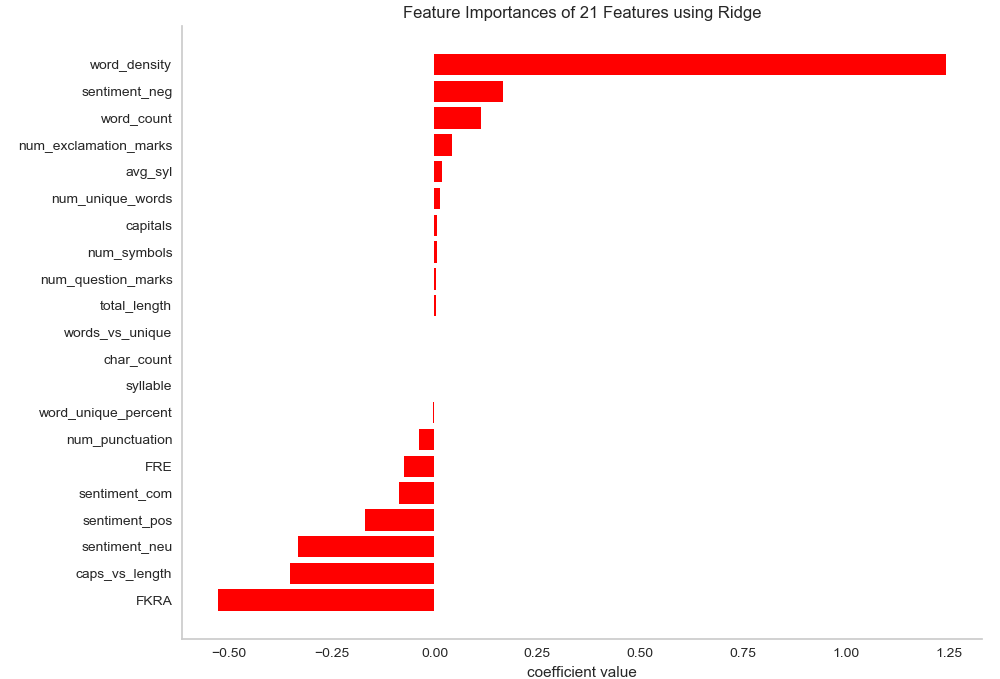}
 
\caption {Feature Importance}
\label{fig:Feature Importance}
\end{figure}

\subsubsection{Text quality: Flesch-Kincaid Reading Ease and Flesch Grade Level Score:}  Readability means how comprehensive content is to understand. Flesch-Kincaid Grade Level and Flesch Reading Ease scores \cite{kincaid1975derivation} are widely accepted matrices to determine the readability of any text content. In both cases, the higher the score, the better it is to understand the content. In hate speech detection on the Twitter data set, these two readability scores are taken as a feature
here \cite{davidson2017automated}. Inspired by that, we have also taken readability as a feature.

\subsubsection{Sentiment scores of tweets} 
VADER (Valence Aware Dictionary and Sentiment Reasoner) is a lexicon and rule-based sentiment analysis tool that is specifically attuned to sentiments expressed in social media. VADER uses a combination of sentiment lexicons, a list of lexical features (e.g., words), which are generally labeled according to their semantic orientation as either positive or negative. VADER not only tells about the positivity and negativity score but also tells us about how positive or negative a sentiment is. For instance,  `I am very sad today.' The sentence is rated as 50.2\% negative, 40.1\% neutral, and 0\% positive.

\subsubsection{Quantitative Features:} 

Apart from the linguistic features, we employ quantitative features such as word count, character count, hashtag count, annotation count, etc. We want to investigate whether such features imply the features of hate or non-hate tweets. A common approach to eliminate features is to describe their relative importance to a model, then eliminate weak features or combinations of features, and re-evaluate to see if the performances are better during cross-validation. To estimate feature importance relative to the model, we need to rank them by feature importance attribute when data is fitted to the model. The Yellowbrick Feature Importances (a scikit learn visualizer tool) visualizer utilizes this attribute to rank and plot features' relative importance. In Fig. \ref{fig:Feature Importance}, the feature importance of the mentioned quantitative features are shown. Some features such as average syllable, capitals, word\_vs\_unique, character\_count, word\_unique\_percentage have no positive or negative correlation with the model.

\subsubsection{Model Architecture and Hyper-parameter Tuning:} 
\begin{figure}[!h]

\minipage{0.32\linewidth}
  \includegraphics[width=\linewidth]{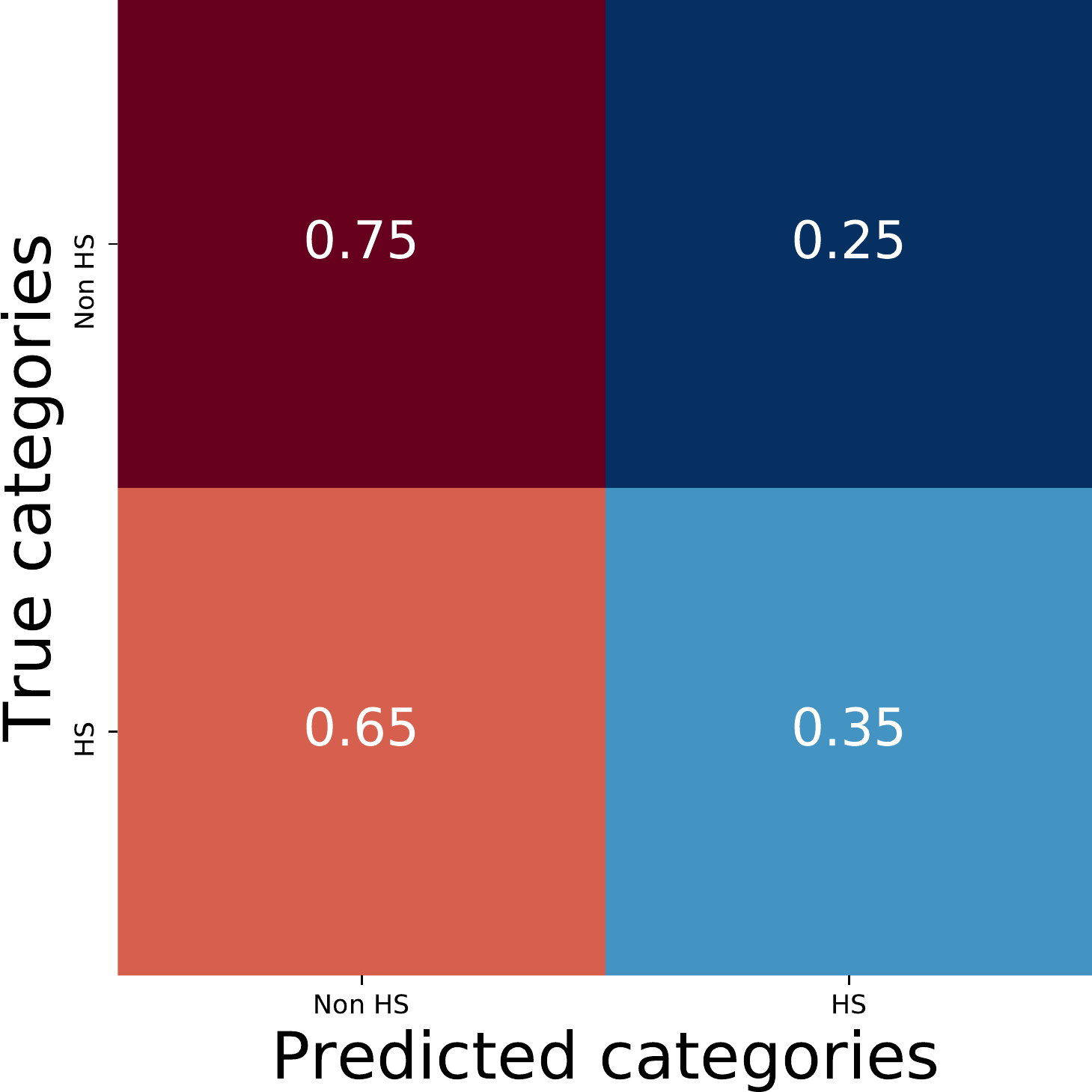}
  \subcaption{Confusion matrix hate speech in English (Linear SVC)}
  \label{fig:Confusion matrix hate speech in English (Linear SVC)}
\endminipage\hfill
\minipage{0.32\linewidth}
  \includegraphics[width=\linewidth ]{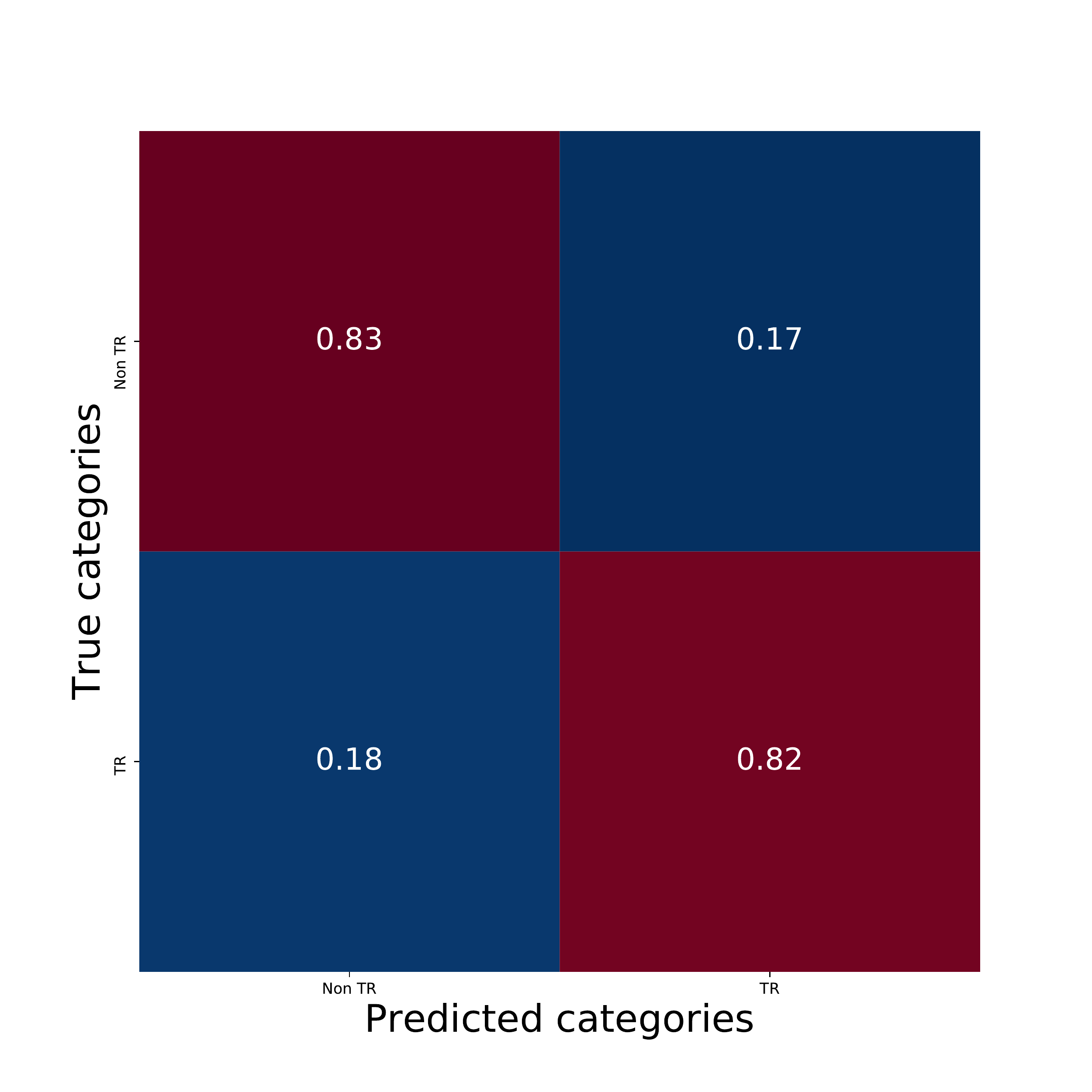}
  \subcaption{Confusion matrix targeted hate speech in English (GRU)}
  \label{fig:Confusion matrix targeted hate speech in English (GRU)}
\endminipage\hfill
\minipage{0.33\linewidth}
  \includegraphics[width=\linewidth ]{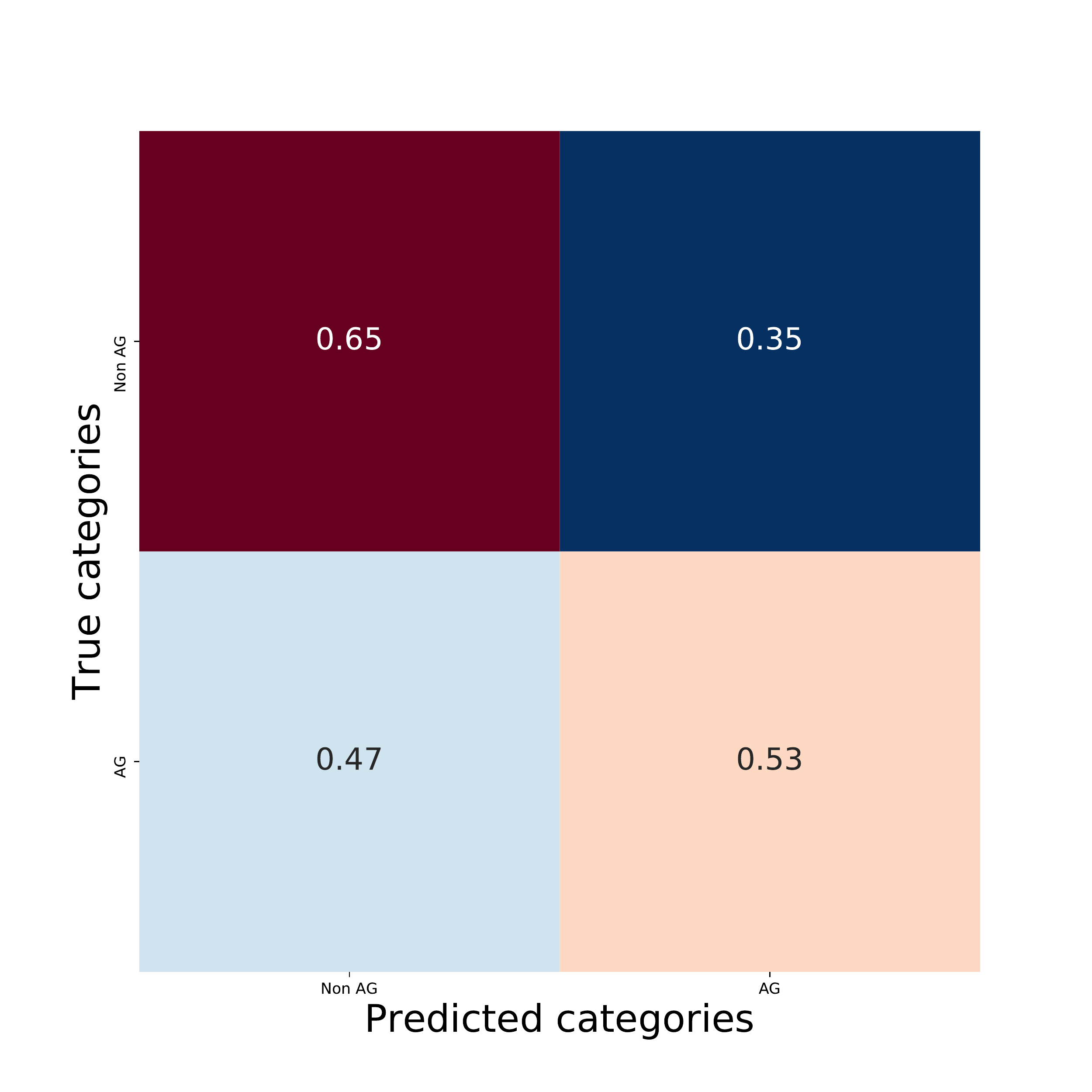}
  \subcaption{Confusion matrix aggressive hate speech in English (GRU)}
  \label{fig:Confusion matrix aggressive hate speech in English (GRU)}
\endminipage

\minipage{0.32\linewidth}
  \includegraphics[width=\linewidth ]{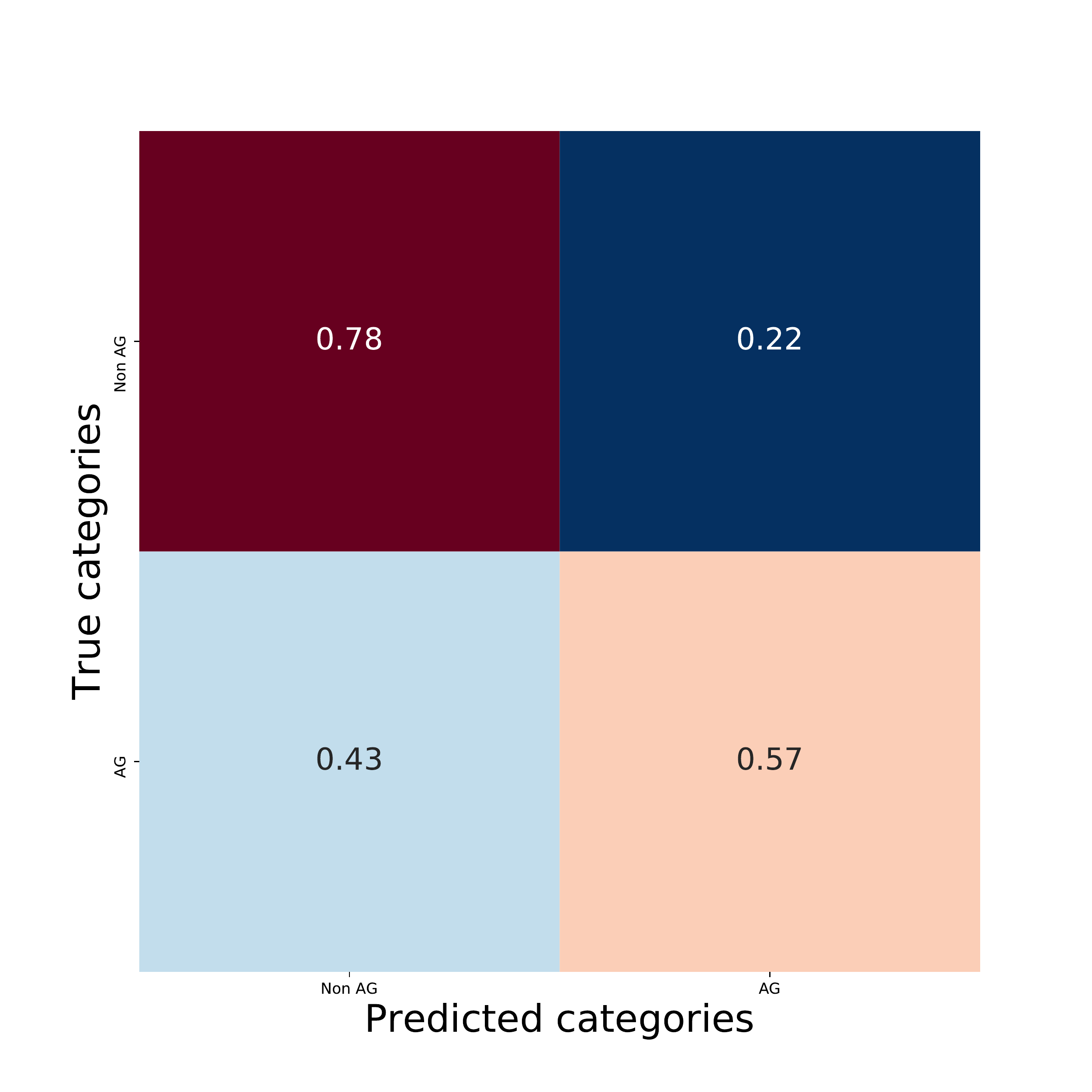}
  \subcaption{Confusion matrix hate speech in Spanish (GRU)}
  \label{fig:Confusion matrix hate speech in Spanish (GRU)}
\endminipage\hfill
\minipage{0.32\linewidth}
  \includegraphics[width=\linewidth ]{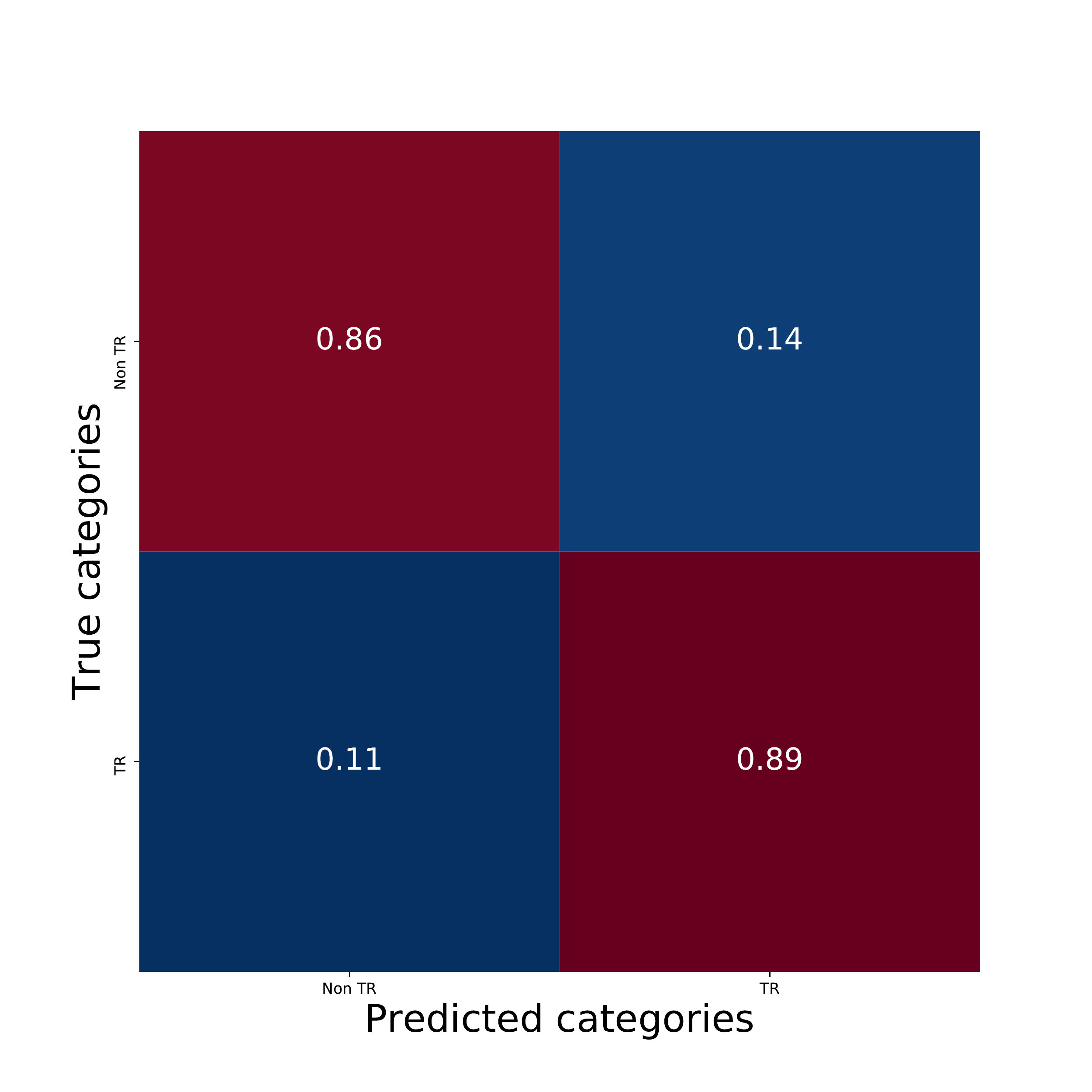}
  \subcaption{Confusion matrix targeted hate speech in Spanish (GRU)}
  \label{fig:Confusion matrix targeted hate speech in Spanish (GRU)}
\endminipage\hfill
\minipage{0.33\linewidth}
  \includegraphics[width=\linewidth ]{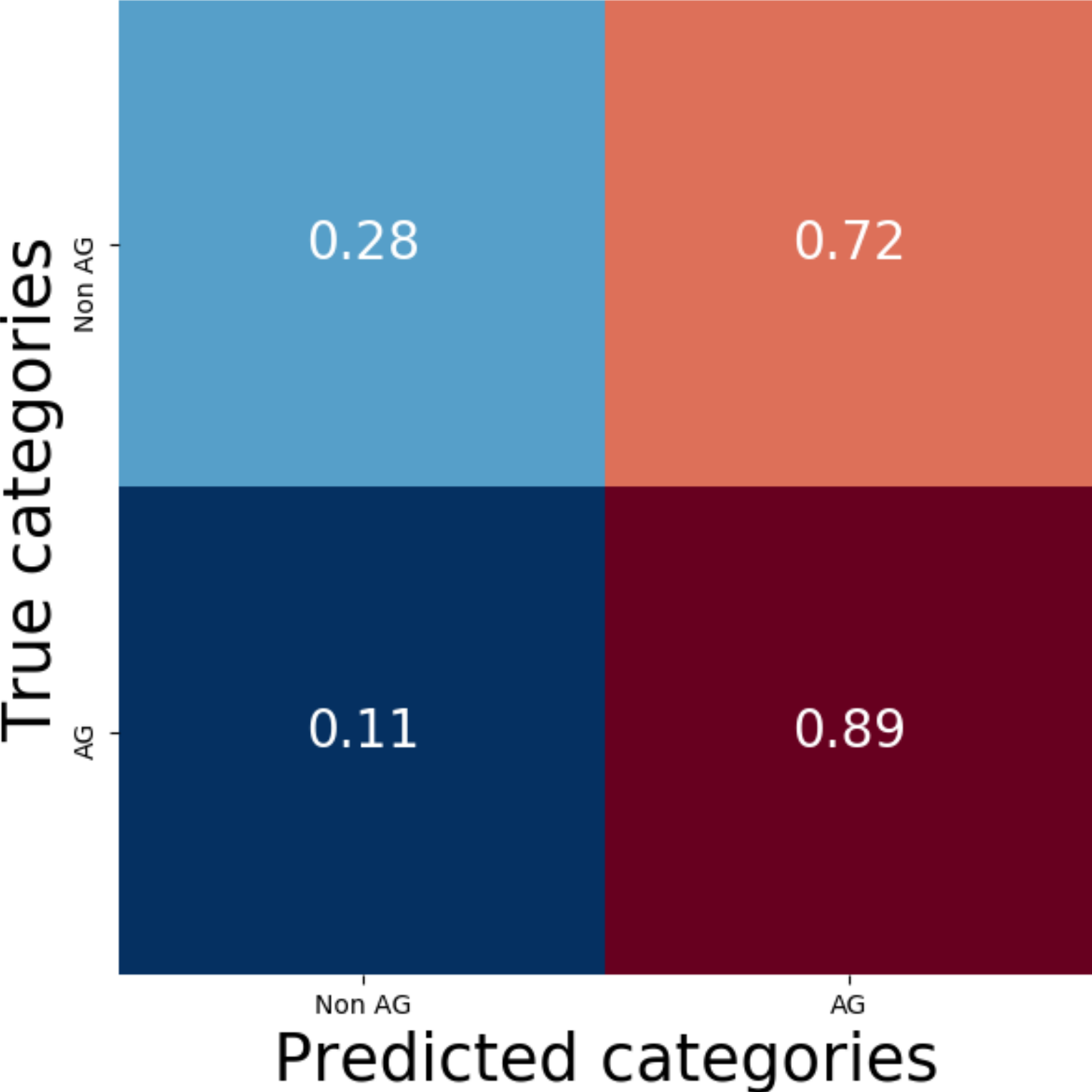}
  \subcaption{Confusion matrix aggressive hate speech in Spanish (GRU)}
  \label{fig:Confusion matrix aggressive hate speech in Spanish (GRU)}
\endminipage

\caption {Confusion Matrix for best models for each task}
\label{fig:Confusion Matrix for best models for each task}
\end{figure}

We first used logistic regression with L1 regularization to reduce the dimensionality of the data. Then, we tested a couple of models like logistic regression, naive bayes, decision trees, random forests, and linear SVC. We exploited each model using 10-fold cross-validation, holding out 10\% of the sample for evaluation to help prevent over-fitting. Af ter using a grid-search to iterate over the models, among all models, linear SVC performed significantly better than other models. After the grid search over the model, the optimal hyper-parameters (C = 0.1, L2 regularization for logistic regression) were fine-tuned. We decided to use logistic regression with L2 regularization for the final model as it more readily allowed us to examine the predicted probabilities of class membership for good performance. We used a one-versus-rest class, and the class label with the highest predicted probability across all classifiers was assigned to each tweet. We applied the scikit-learn implementation for the proposed model. In Table \ref{tab:N gram feature weights}, the top most N- gram feature weights are shown. In Table \ref{tab:Macro average f1 score for two subtask in Linear SVC}, the macro-average F1-score is shown. The score improves significantly than earlier models (GRU and CNN) in the case of hate speech detection in the English language, which is the primary concern of this proposed model. In Table \ref{tab:Comparison between the current best models and ours} the current best scores (among the participants of the competition) are compared with the proposed models. In Table \ref{tab:Performance comparison between subtask A and subtask B}, the performance of the subtask A and subtask B are presented. In the case of all subtasks except subtask A (English),  for subtask B (English, Spanish) the score is improved 10, 5 points respectively. In Fig. \ref{fig:Confusion Matrix for best models for each task}, the confusion matrix of the best models are shown.

\begin{table}{}
\centering
\caption{Macro average f1 score for two subtask in Linear SVC}
\label{tab:Macro average f1 score for two subtask in Linear SVC}

\begin{tabular}{ |c| c | }
\hline
 Class & F1-score\\
 \hline
 EN\_HS &  57\\ 
 \hline
 EN\_TR &  79 \\  
 \hline
 EN\_AG &  51 \\   
 \hline
 ES\_HS &  52\\ 
 \hline
 ES\_TR &  80 \\  
 \hline
 ES\_AG &  53 \\
 \hline
\end{tabular}
\end{table}
\begin{table}
      \centering
        \caption{N gram feature weights}
        \label{tab:N gram feature weights}
\begin{tabular}{|c|c|}
\hline
N gram & weight\\
\hline
women are stupid &           0.117489 \\
\hline
bitch https co    &          0.075791  \\
\hline
build the wall     &         0.075195 \\
\hline
trump maga rednationrising & 0.072382 \\
\hline
are not welcome   &          0.065893 \\
\hline
most hysterical woman &      0.060286  \\
\hline
shut the fuck    &      0.048063  \\
\hline
\end{tabular}{}
\end{table}

\begin{table}{}


      \centering
              \caption{Performance comparison between the current best models and proposed model (in terms of macro average F1 score).}
\label{tab:Comparison between the current best models and ours}
\begin{tabular}{ |c| c| c|c|}
\hline
 Class & best F1 & new F1 & model\\
 \hline
 EN\_HS &  65  & 57 & Linear SVC\\ 
 \hline
 EN\_TR &  - & 83 & GRU\\  
 \hline
 EN\_AG & - &  61 & GRU\\   
 \hline
 ES\_HS &  73 & 75 & GRU\\ 
 \hline
 ES\_TR & - & 88 & GRU\\  
 \hline
 ES\_AG & - & 65 & GRU\\
 \hline
\end{tabular}
\end{table}
\begin{table}{}
      \centering  
            \caption{Performance comparison between subtask A and subtask B (in terms of macro average F1 score).}
\label{tab:Performance comparison between subtask A and subtask B}
\begin{tabular}{ |c|c|c|c|}
\hline
 Tasks & Old best F1 & New F1 \\
 \hline
  Subtask-A (ENG) &  65  & 57 (second best)\\ 
 \hline
 Subtask-A (ESP) &  73 & 75 \\  
 \hline
 Subtask-B (ENG) & 57 &  67 \\   
 \hline
 Subtask-B (ESP) & 70.5   & 75.33\\ 
 
 \hline
\end{tabular}
\end{table}

\section{Dataset Analysis and Challenges for Hate Speech Detection in English}

\begin{figure}[h!]
    \centering
    \includegraphics[width=\linewidth]{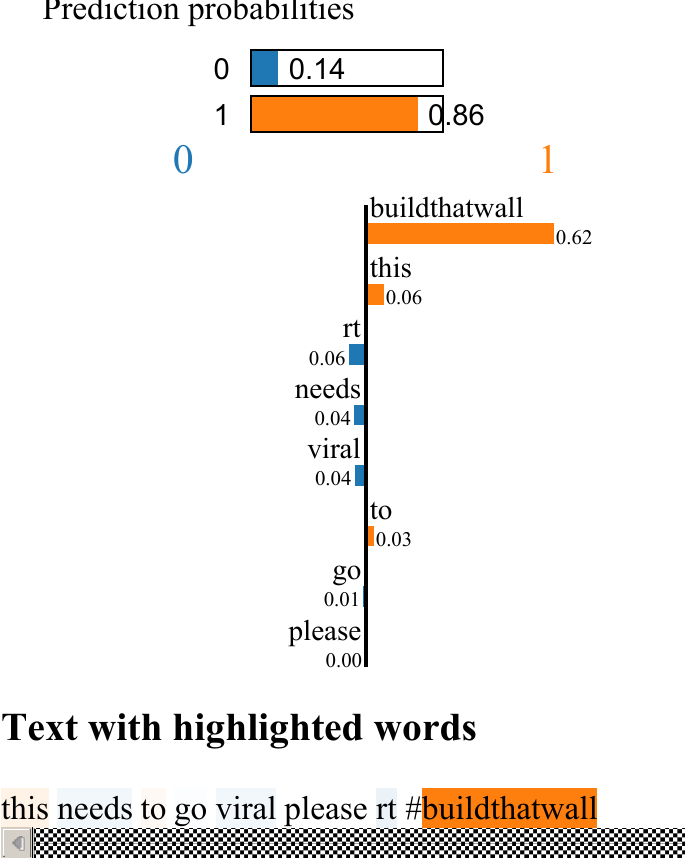}
    \caption{LIME explanation}
    \label{fig:LIME explanation}
\end{figure}

From the discussion of previous sections, to develop an efficient model for hate speech detection (subtask A) in the English language is more challenging than the rest of the tasks. Some quantitative investigations can shed some light here. The topmost hashtags in both train and test tweets are \textit{\#BuildTheWall}, \textit{\#BuildThatWall}, \textit{\#MAGA}, \textit{\#TRUMP}, \textit{\#NODACA}, \textit{\#WOMENSUCK}, \textit{\#IMMEGRANT}, etc. Among those, \textit{\#BuildTheWall} or \textbf{\#BuildThatWall} are used as 9\% and 12\% of the total hashtags in train and test dataset respectively. \textit{\#MAGA} also contributes to the 3\% or 6\% of the total hashtags in train and test dataset respectively. However, while tweets mentioning \textit{\#BuildThatWall} or \textit{\#BuildTheWall} were annotated as hate speech 98\% of the time in the training set, this number is just 35\% on the test set. Similarly, tweets containing \textit{bitch} were labeled as hate speech 78\% of the time in the training set vs. 43\% of the time in the test set. Therefore, the discrepancy exhibited in the official training and test dataset, especially for English subtask A, makes it difficult for improving the F1-score for hate speech detection. In table \ref{tab:Non hate speech treated as hate speech in test set}, we present misclassified examples of hate tweets that are non hate according to annotation. However, similar types of tweets are annotated as hate in the train set. In Fig. \ref{fig:LIME explanation}, we have shown which features are used in linear SVC model to classify the hate or non-hate using LIME \cite{LIME}.  Here, the prediction probability of hate speech is 86\%, and if the hashtag is removed the probability is reduced by 62\%. 
 
 \begin{table}{}         
      \centering  
            \caption{Non hate speech classified as hate speech in test set}
\label{tab:Non hate speech treated as hate speech in test set}
\begin{tabular}{ |c|}
\hline
  who love illegal aliens you all have more blood on your \\ hands you have betrayed america \#illegalimmigrants \\
 \hline
  just as scuzball democrats want need \#buildthewall \#sendthemhome\\
  \hline
  this needs to go viral please rt \#buildthatwall \\
  \hline
\end{tabular}
\end{table}

Now, we present some interesting cases where the system misclassifies the non hate speech as hate speech and vise versa. For example: \\

(1) \textit{ \textbf{bitch} you lying to your damn self thinking youre \textbf{ugly} httpstcouyipjkgx} (Actual Non hate, predict hate)

(2) \textit{ i wasnt talking about you in my tweet but if that shoe fits ya then lace that \textbf {bitch} up and wear it} (Actual hate, predict non hate )

These are the two misclassified examples of tweets of the linear SVC model. In both tweets having some offensive words, however, it is unclear at which context the first one is labeled as non-hate speech. In the case of the first one, according to LIME, the most important features to be considered as hate are \textit{bitch} and \textit{ugly}. In the second one, the probability is 51\% to be non-hate, and the pattern of the sentence is a bit sloppy and ambiguous.

On the other hand, the use of hashtags differs markedly between languages. Hashtags are more frequent in English training data than Spanish training data. In English tweets, it is 2.6 times more likely to contain at least one tag, and with more than one hashtag, it is 4.1 times the rate in Spanish. In the English training data, the most frequent ten hashtags were 23\% of the overall total and tended towards American political topics. In Spanish, the top ten hashtags account for only 8\% of the total, exhibiting a much longer and sparser tail. The discrepancy exhibited between the train and test dataset in the English language hurts the F1-score for hate speech identification task.

\section{Error Analysis }

Error analysis was carried out to analyze the misclassifications that we encountered in our system by quantitative analysis using the confusion matrix of our best models for each task. 

\subsection{Quantitative Error Analysis:}
From Fig. \ref{fig:Confusion Matrix for best models for each task}, we can compare the true positive and false negative rates of each task. The false-positive rate for a hate speech class in English is around 65 \%. In the previous section, we have discussed why the Linear SVC model has a larger error rate in English hate speech detection by presenting the discrepancies in the dataset as well as annotations task. On the other hand, the targeted hate speech detection for both English and Spanish has a high true positive and low false-positive rate. From Fig. \ref{fig:Confusion matrix targeted hate speech in English (GRU)}, \ref{fig:Confusion matrix targeted hate speech in Spanish (GRU)} the false positive rates are  18\% and 11\%  for targeted hate speech in English and Spanish. In terms of hate speech against immigrants and women, the false positive rate is 47\%. In the previous section, we provided statistical analysis of how different words (e.g., bitch, ugly) are used for annotation of hate and non-hate content differently in train and test sets. Therefore, the model trained over the features for hateful content towards women fails to correctly classify in test sets due to different annotation strategies. 

\subsection{problematic labelled dataset:} 
We have discussed how the discrepancy between the annotation task of train and test set can impact the model performance. To address this issue, Chen et al \cite{Chen} proposed an aggregated mechanism where rather than a binary classification scheme, annotators have \textit{undecided} option for ambiguous content. As a result, the classifier performance improved by 18\% based on syntactic, semantic, and context-based features.

\section{Conclusion}

Hateful speech detection has become challenging because of the diversity of the languages and usages pattern of the users. In this paper, some state-of-the-art machine learning or deep learning algorithms with textual feature engineering by investigating linguistic features were employed to classify tweets according to whether they contain hate speech, aggression, and targeting hate speech towards women or individuals. The novel contribution includes the improvement in the performance matrices (F1 score) among the existing participants by proposing new models. Moreover, the quantitative analysis on the dataset presents the discrepancy between the official train and test dataset, particularly in the English subtasks, and describes how this impacts the performances of the model.

\bibliographystyle{IEEEtran}
\bibliography{sample}

\end{document}